\documentclass[runningheads]{llncs}
\usepackage{graphicx}

\usepackage{tikz}
\usepackage{comment}
\usepackage{amsmath,amssymb} 
\usepackage{color}
\usepackage{booktabs}
\usepackage{multicol}
\usepackage{multirow}
\usepackage{wrapfig}
\usepackage{colortbl}
\usepackage{xcolor}         
\usepackage{color}
\usepackage{float}
\definecolor{mygray}{gray}{.9}
\definecolor{mycyan}{cmyk}{.3,0,0,0}
\definecolor{light-gray}{gray}{0.5}
\definecolor{c0}{RGB}{146,208,80}
\definecolor{c1}{RGB}{0,176,80}
\definecolor{c2}{RGB}{102,153,0} 
\definecolor{c3}{RGB}{0,0,255}
\definecolor{ACM-purple}{RGB}{121,29,125}
\usepackage[accsupp]{axessibility}  

\newcommand{\tabincell}[2]{\begin{tabular}{@{}#1@{}}#2\end{tabular}}

\begin{document}
\pagestyle{headings}
\mainmatter

\title{Region-aware Knowledge Distillation for Efficient  Image-to-Image Translation} 

\author{Linfeng Zhang$^1$, Xin Chen$^2$, Runpei Dong$^3$, Kaisheng Ma$^1$\thanks{Corresponding author.}}
\institute{Tsinghua University$^1$, Intel Corporation$^2$, Xi'an Jiaotong University$^3$}

\maketitle
\begin{abstract}  
  Recent progress in image-to-image translation has witnessed the success of generative adversarial networks (GANs). However, GANs usually contain a huge number of parameters, which lead to intolerant memory and computation consumption and limit their deployment on edge devices. To address this issue, knowledge distillation is proposed to transfer the knowledge from a cumbersome teacher model to an efficient student model. However, most previous knowledge distillation methods are designed for image classification and lead to limited performance in image-to-image translation.
  In this paper, we propose Region-aware Knowledge Distillation (\texttt{ReKo}) to compress image-to-image translation models. 
  Firstly, \texttt{ReKo} adaptively finds the crucial regions in the images with an attention module. Then, patch-wise contrastive learning is adopted to maximize the mutual information between students and teachers in these crucial regions. Experiments with eight comparison methods on nine datasets demonstrate the substantial effectiveness of \texttt{ReKo} on both paired and unpaired image-to-image translation. For instance, our 7.08$\times$ compressed and 6.80$\times$ accelerated CycleGAN student outperforms its teacher by 1.33 and 1.04 FID scores on Horse$\rightarrow$Zebra and Zebra$\rightarrow$Horse, respectively. Codes will be released  on GitHub.
  \end{abstract}

  \section{Introduction}
  Tremendous breakthroughs have been attained with the state-of-the-art generative adversarial network (GAN) in generating high-resolution, high-fidelity, and photo-realistic images and videos~\cite{singan,bigGAN,gan,pix2pix,cyclegan,DBLP:conf/cvpr/ZhaoC21}.
  Because of its powerful ability of representation and generation, GAN has evolved to the most dominant model in image-to-image translation~\cite{new_style1,new_style2,new_style3,new_style4,DBLP:conf/cvpr/ChandranZG0B21,DBLP:conf/cvpr/KotovenkoWHO21}. However, the advanced performance of GAN is always accompanied by tremendous parameters and computation, which have limited their usage in resource-limited edge devices such as mobile phones. 
  
  To address this issue, knowledge distillation is proposed to improve the performance of an efficient student model by mimicking the features and prediction of a cumbersome teacher model. Following previous research on image classification~\cite{fitnets,relational_kd2}, some recent works have tried to directly apply knowledge distillation to image-to-image translation but earned very limited improvements~\cite{gan_compress,spkd_gan}.
  In this paper, we first argue that most previous knowledge distillation methods ignore the \emph{spatial redundancy} in image-to-image translation, which results in their failure.
  More specifically, in image-to-image translation, only a few regions of the images are actually required to be translated usually. For example, in the Horse$\rightarrow$Zebra task, only the regions of horses need to be translated while the regions of background should be preserved. Even in some tasks where the whole image is required to be translated, usually there are relatively some more crucial regions. However, previous knowledge distillation methods directly employ the student to mimic teacher features in all the regions with the same priority while ignoring the spatial redundancy.
  
  Since  students have much fewer parameters than their teachers, usually they are not able to learn all the knowledge from teachers. Thus, in knowledge distillation, the student should pay more attention to teacher knowledge in the crucial regions instead of learning all the regions with the same priority. Unfortunately, different from the other vision tasks such as object detection, there are no annotations on crucial regions in image-to-image translation, especially unpaired image-to-image translation. Thus, it is still challenging to localize and make good use of these crucial regions. 
  To tackle this challenge,  in this paper, we propose a novel knowledge distillation method referred to as Region-aware Knowledge Distillation (\texttt{ReKo}), which mainly contains the following two steps.
  \begin{figure}[t!]
    \begin{center}
        \includegraphics[width=11.5cm]{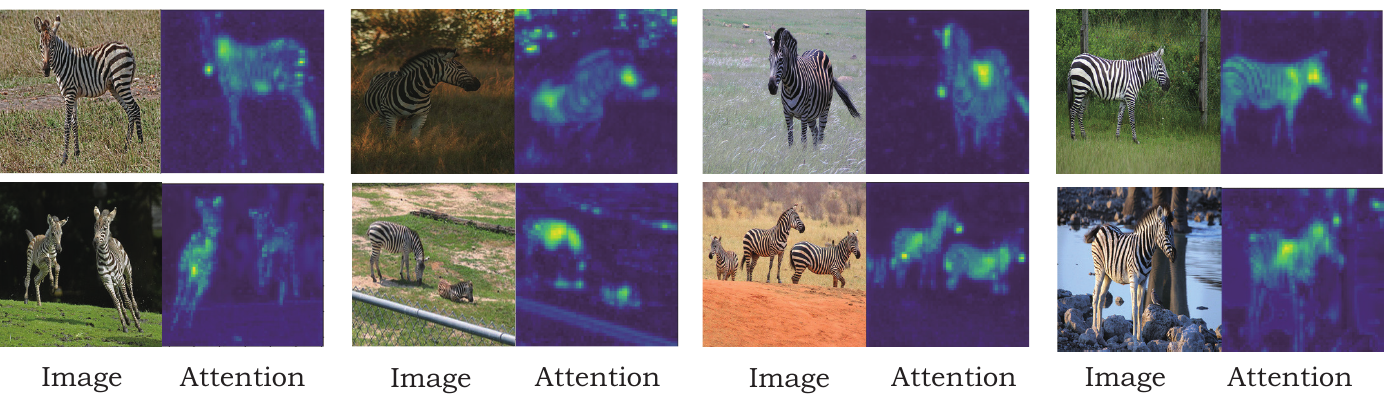}
        \vspace{-0.4cm}
    \end{center}
    \caption{\label{fig:figure1} Visualization on the attention results on Zebra$\rightarrow$Horse. The attention module in \texttt{ReKo} can localize the to-be-translated objects (zebras) unsupervisedly.}
    \vspace{-0.25cm}
\end{figure}

  Firstly, \texttt{ReKo} localizes the crucial regions in an image with a parameter-free attention module and then only distills teacher features in these crucial regions.
  As discussed in previous works~\cite{attention_zhoubolei,detectiondistillation,attentiondistillation}, the attention value in a region shows its importance. A region with higher attention value usually has more influence on the prediction of the neural network, and thus should be considered as a more important region.
  Hence, we define the importance of a region as its attention value, which is further utilized to decide whether teacher features in this region should be distilled to the student. Visualization results of this attention module have shown in Figure~\ref{fig:figure1}. It is observed that this method successfully localizes the regions of horses while filtering the regions of background.
  
  Secondly, \texttt{ReKo} adopts a patch-wise contrastive learning framework to optimize knowledge distillation. 
  Instead of distilling teacher knowledge to students by directly minimizing the $L_2$-norm distance between their features, we propose to adopt a contrastive learning framework for optimization.
  Tian~\emph{et al.} firstly show that on image classification, knowledge distillation can be performed with contrastive learning to maximize the mutual information between students and teachers~\cite{kd_crd}.
  However, their method requires a huge memory bank to contain massive negative samples\footnote{16384 negative samples are required according to their released codes.}, which is not applicable for image-to-image translation. To address this issue, we propose to apply patch-wise contrastive learning framework~\cite{park2020contrastive} for knowledge distillation, which regards student and teacher features in the same patch as a positive pair and the other features as negative pairs. During the distillation period, by optimizing these pairs with InfoNCE loss~\cite{cpc}, the similarity between student and teacher features in the same region is improved, and thus teacher knowledge is distilled to the student.

\begin{figure*}[t]
    \centering
    \includegraphics[width=12cm]{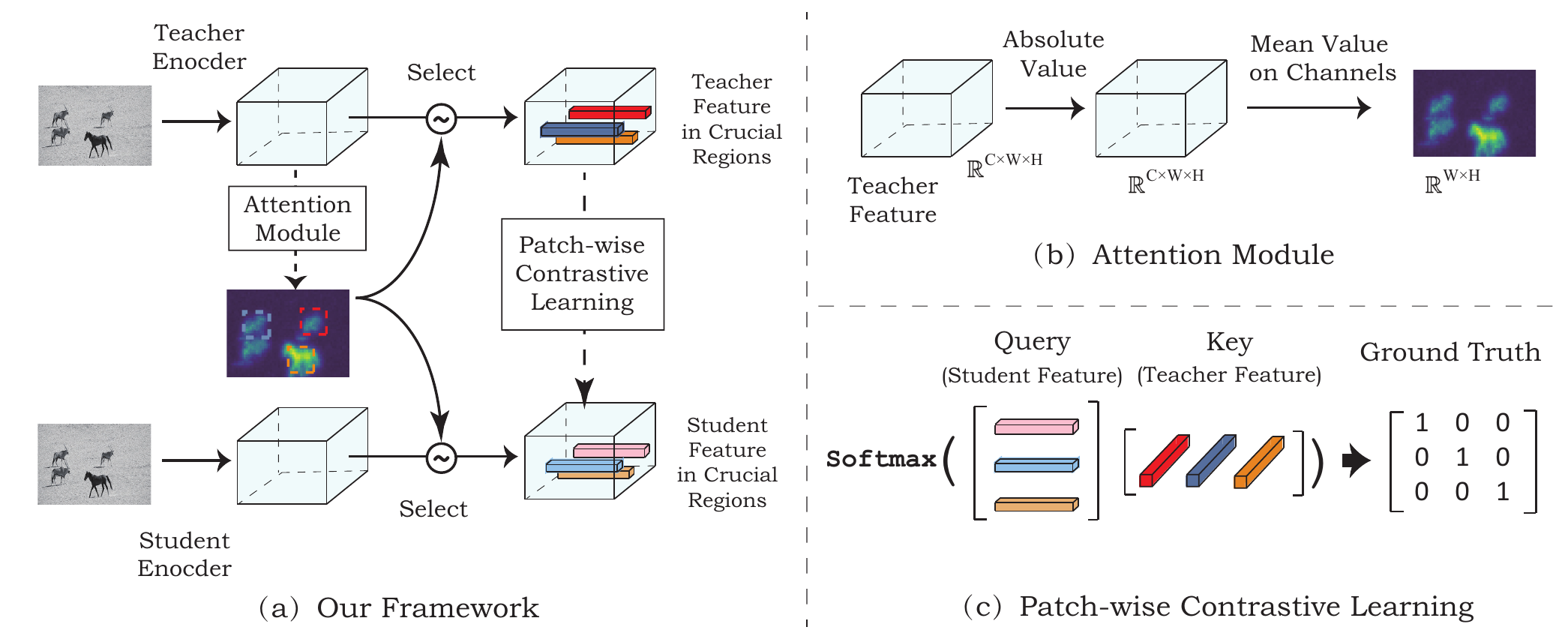}
    \caption{\label{fig:region_aware} The overview of Region-aware Knowledge Distillation (best viewed in color). It mainly consist of two steps. \textbf{ Step-1}: Find the crucial regions in the to be translated image by applying the \texttt{attention module} to teacher features. Note that the attention module is composed of an absolute value operation and a mean operation on the channel dimension. Then, $K$ regions with the $K$ largest attention values are selected as the crucial regions (here $K$=3). \textbf{Step-2}: Perform knowledge distillation in these crucial regions with \texttt{patch-wise contrastive learning}.
    Student features and teacher features in the same region (such as \textcolor{pink}{$\blacksquare$} and \textcolor{red}{$\blacksquare$} ) are considered as a positive pair and the others (such as {\textcolor[RGB]{31,163,221}{$\blacksquare$} } and \textcolor{red}{$\blacksquare$}) are regarded as negative pairs. All these pairs are optimized in a contrastive learning framework with InfoNCE loss, which regards the student features as queries and teacher features as the keys.  }
    \vspace{-0.3cm}
\end{figure*}
  
  Extensive experiments on nine datasets with eight comparison methods have been conducted to demonstrate the effectiveness of \texttt{ReKo} both quantitatively and qualitatively. The visualization results between students and teachers show that \texttt{ReKo} has successfully increased the feature similarity between students and teachers in the same region, which indicates knowledge has been effectively distilled. Moreover, we show that \texttt{ReKo} is able to stabilize the training of GANs and thus prevent them from model collapse. To sum up, our main contributions in this paper can be summarized as follows.
  
  \begin{itemize}
  \item We propose \emph{Region-aware Knowledge Distillation} (\texttt{ReKo}), which exploits the spatial redundancy in image-to-image translation for model compression. Instead of distilling teacher features in all the regions, \texttt{ReKo} first adaptively localizes the crucial regions in the image with an attention module and then only performs knowledge distillation in these crucial regions. 
  \item We propose to transfer teacher knowledge in the crucial regions with \emph{patch-wise contrastive learning}. By improving the mutual information between student features and teacher features in the same patch, it significantly improves the effectiveness of knowledge distillation without requirements on additional memory banks and a large batch size.
      
  \item Experimental results with eight comparison methods on nine datasets have demonstrated the effectiveness of \texttt{ReKo} on both paired and unpaired image-to-image translation. For instance, our 7.08$\times$ compressed and 6.80$\times$ accelerated CycleGAN student outperforms its teacher by 1.33 and 1.04 FID scores on Horse$\rightarrow$Zebra and Zebra$\rightarrow$Horse, respectively.
     
  \end{itemize}

  \section{Related Work}
  \subsection{GANs for Image-to-Image Translation}
  Generative Adversarial Network (GAN), which is composed of a generator for image generation and a discriminator for discriminating the real and generated images, have become the most popular model in image-to-image translation~\cite{gan}.
  Pix2Pix is proposed to apply conditional GAN~\cite{mirza2014conditional} to  image-to-image translation on paired datasets~\cite{pix2pix}.
  Then, Pix2PixHD improves the resolution of generated images with multi-scale neural networks and boundary maps~\cite{pix2pixHD}.
  Based on these efforts, Wang \emph{et al.} further propose Vid2Vid to perform video-to-video translation~\cite{vid2vid}. Recently, GANs have been utilized in various image-to-image translation tasks such as single image super resolution~\cite{srgan,wang2018esrgan}, image deblurring~\cite{kupyn2018deblurgan} and image style transfer~\cite{stylegan,styleganv2,DBLP:conf/cvpr/0002H0HS21,DBLP:conf/cvpr/RichardsonAPNAS21,DBLP:conf/cvpr/LiuSCBZSL0N21,DBLP:conf/cvpr/LiZ0CHMHWJ21,DBLP:conf/cvpr/KimKC21}. A more challenging task in this domain is how to perform image-to-image translation on unpaired datasets.
  CycleGAN, DualGAN, and DiscoGAN have been proposed to solve this issue by regularizing the training of generators with the cycle consistency loss, which aims to minimize the difference between the origin images and the back translation images ~\cite{cyclegan,dualgan,discogan}.
  Recently, Park~\emph{et~al.} propose to replace the cycle consistency loss with PatchNCE loss, which minimizes the mutual information between the corresponding input and output patches and thus does not require bi-directional generators~\cite{park2020contrastive}.

  \subsubsection{Attention Methods in GANs} Recently, abundant works have been proposed to improve the performance of GANs with attention methods. Usually, Attention GANs are composed of a transformation network
  which actually carries out the object transformation and an attention network (module) to predict the spatial location of to-be-transformed objects~\cite{attentiongan1,attentiongan2,attentiongan3,attentiongan5,attentiongan6,attentiongan4}. Based on the prediction from attention networks, the transformation network can focus more on the relatively more important regions, and thus achieve better performance than previous methods.
  Although attention modules are also utilized in our method, it has essential difference as follows: (i) Attention layers in previous methods are utilized as a build-in module in GANs and required in both training and inference.
  In contrast, the attention layer in \texttt{ReKo} is only utilized during training and thus does not introduce any computational and storage cost in inference. (ii) Most previous Attention GANs are proposed to improve the quality of generated images while \texttt{ReKo} is a knowledge distillation method for model compression. (iii) The attention layers in previous methods are layers with parameters which are required to be trained with the model. In contrast, the attention layer in \texttt{ReKo} is parameter-free, which does not require any specific training.
  Moreover, the usage of the attention layer in our method is an effective but not the only solution to find the crucial regions in an image. It can be replaced by the other methods such as intergrated gradients~\cite{sundararajan2017axiomatic}, saliency detection~\cite{DBLP:journals/pami/IttiKN98}, and Grad-CAM~\cite{DBLP:journals/ijcv/SelvarajuCDVPB20} which have been  introduced in Section~\ref{sec:other}.
 
 \subsubsection{GAN Compression} The tremendous storage and computation consumption in GAN have promoted the research on its compression.
  Wang~\emph{et al.} propose a unified GAN compression framework with knowledge distillation, channel pruning, and quantization~\cite{wang2020gan}. Li~\emph{et al.} propose to compress GANs with once-for-all neural network architecture search and na\"ive feature knowledge distillation~\cite{gan_compress}.
  Shu~\emph{et al.} propose to investigate and prune the unimportant weights in GANs with a  co-evolutionary approach~\cite{shu2019co}.
  Recently, Jin~\emph{et al.} introduce an inception residual block into generators and prune it with a one-step pruning algorithm~\cite{teacher_do_more_gankd}.
  Li~\emph{et al.} propose a generator-discriminator cooperative compression scheme, which applis selective activation discriminators to maintain the Nash equilibrium in adversarial training to avoid model collapse~\cite{revisit_discriminator}.

  \subsection{Knowledge Distillation}
  Knowledge distillation has become one of the most effective techniques for model compression~\cite{model_compression,distill_hinton}. It first trains a cumbersome teacher model and then transfers its knowledge to a lightweight student model. Previous knowledge distillation usually aims to distill the knowledge in the logits (softmax outputs)~\cite{distill_hinton,deepmutuallearning}. Then, abundant methods have been proposed to distill the knowledge in the features and its variants, such as attention~\cite{attentiondistillation,detectiondistillation} and the gram matrix~\cite{fsp_kd}. Recently, some research has been proposed to distill the relation between different samples~\cite{relational_kd,relational_kd2} and pixels~\cite{structuredkd,local_kd}. 
  A recent popular trend in knowledge distillation is to maximize the mutual information between students and teachers with contrastive learning~\cite{DBLP:conf/cvpr/ZhuTCYLRYW21,DBLP:conf/cvpr/WangHLXY021}. Tian~\emph{et~al.} first propose the contrastive representation distillation framework, which regards the representation of the same image from students and teachers as a positive pair in contrastive learning. Then Chen~\emph{et al.} extend this idea with the Wasserstein distance~\cite{chen2020wasserstein}. However, their methods usually requires a large number of negative samples for optimization, which is not practicable for image-to-image translation.
  In this paper, we extend this framework with patch-wise contrastive learning~\cite{park2020contrastive} for knowledge distillation on image-to-image translation to address this issue. Given an image. patch-wise contrastive learning regards the the feature of different patches as the negative samples, which is efficient in both computation and memory.
  
  \subsubsection{GAN Knowledge Distillation} In the last several years, there has been some research proposed to apply knowledge distillation to the compression of GANs. Li~\emph{et al.} propose to improve the performance of student generators with the na\"ive feature distillation~\cite{gan_compress}. Then, Li~\emph{et al.} propose the semantic relation preserving knowledge distillation, which computes and distills the relation between different patches in generators~\cite{spkd_gan}.  Jin~\emph{et al.} propose to distill the knowledge in features with global kernel alignment, which enables knowledge distillation without additional adaptation layers~\cite{teacher_do_more_gankd}.
  Recently, Liu~\emph{et al.} propose to compress unconditional GANs by parsing  the contents in the images~\cite{liu2021content}. Zhang~\emph{et~al.} propose  wavelet knowledge distillation which decomposes images into different frequency bands and then only distill the high frequency information~\cite{wkd}. Ren~\emph{et al.} introduce online multi-granularity distillation which trains a discriminator-free student in a one-stage distillation scheme~\cite{omgd}. Li~\emph{et al.} propose to distill teacher knowledge in both generators and discriminators with texture loss~\cite{revisit_discriminator}.
  

    \vspace{-0.1cm}
  \section{Methodology}
    \vspace{-0.cm}
  \subsection{Formulation}
    \vspace{-0.0cm}
  \subsubsection{Patch-wise Contrastive Learning   for Knowledge Distillation}
  Given two sets of images $\mathcal{X}$ and  $\mathcal{Y}$, image-to-image translation aims to find a mapping function $\mathcal{F}:\mathbb{R}^{C\times H\times W}\rightarrow\mathbb{R}^{C\times H\times W}$ which maps images in $\mathcal{X}$ to $\mathcal{Y}$. Note that $C, H, W$ indicate the number of channels, height and width of the image, respectively. Usually, $\mathcal{F}$ can be divided into an encoder $\mathcal{F}_{\text{enc}}$ followed by a decoder $\mathcal{F}_{\text{dec}}$. Given an image $x$, then its intermediate feature can be formulated as $\mathcal{F}_{\text{enc}}(x)\in \mathbb{R}^{c \times w \times h}$ where $c$,  $w$ and $h$ denote its number of channels, width and height respectively. For convenience, we reshape it into $\mathbb{R}^{c \times wh}$, where $\mathcal{F}_{\text{enc}}(x)[:,i]$ indicates the feature of $i$-th region. The corresponding index set of regions can be formulated as $S=\{1,2,3,..., wh\}$. In this paper, we adopt a noise contrastive estimation framework~\cite{cpc} to maximize the mutual information between the features of students and teachers. Given a query $v$, a positive key $v^+$ and a set of  negative keys  $\{v^-_1, v^-_2, ..., v^-_N\}$. The InfoNCE loss can be formulated as \vspace{-0.3cm}
  
  \begin{equation}
  \label{equ:contrast_learning}
  \begin{aligned}
  &\mathcal{L}_{\text{InfoNCE}}(v,v^{+},v^{-})=-\log\left[ 
  \frac{
  \exp(v\cdot v^{+}/\tau)}
  {
  \exp(v\cdot v^{+}/\tau)+
  \sum_{n=1}^N \exp(v \cdot v^-_i / \tau)
  } \right],
  \end{aligned}
  \end{equation}
  where $\tau$ is a temperature hyper-parameter.
  By regarding the features of students and teachers at the same region (patch) as positive pairs and the other features as the negative pairs, we can extend InforNCE to patch-wise contrastive distillation framework, which can be formulated as 
  \begin{equation}
  \begin{aligned}
      &\mathcal{L}_{\text{RegionDis}}(\mathcal{X},  \mathcal{F}_{\text{enc}}^\mathcal{S}, \mathcal{F}_{\text{enc}}^\mathcal{T}) \\&= \mathbb{E}_{x \sim \mathcal{X}}\mathop{\sum}^{wh}_{i=1} \mathcal{L}_{\text{InfoNCE}}(
      \underbrace{\mathcal{F}^{\mathcal{S}}_{\text{enc}}(x)[:,i]}_{\text{Query $(v)$}}
      \underbrace{\mathcal{F}^{\mathcal{T}}_{\text{enc}}(x)[:,i]}_{\text{Positive Key $(v^+)$}},~
     \underbrace{ \{ \mathcal{F}_{\text{enc}}^\mathcal{T}[:,j]~|~ j\in S, j\neq i \})}_{\text{Negative Keys $(v^-)$}},
      \label{RegionDis}
      \end{aligned}
  \end{equation}
  where the scripts $\mathcal{S}$ and  $\mathcal{T}$ are utilized to distinguish student networks and teacher networks, respectively. During knowledge distillation, the distance between the query and the positive key (student and teacher features in the same region) are minimized 
  and the distance between the query and the negative queries (student and teacher feature in different regions) are maximized. Thus, teacher knowledge is distilled to the student.
  
\vspace{-0.25cm}

    \subsubsection{Parameter-free Attention Module} It is generally acknowledged that the attention value of each pixel shows its importance~\cite{detectiondistillation}. In this paper, we define the attention value of a region as its absolute mean value across the channel dimension, which can be formulated as 
  \begin{equation}
  \mathcal{A}: \mathbb{R}^{c\times wh}\xrightarrow[]{\text{absolute}}\mathbb{R}^{c\times wh}\xrightarrow[]{\text{mean on channel}}\mathbb{R}^{ wh}.
  \end{equation}
  \vspace{-0.65cm}
    \subsubsection{Distilling Only the Crucial Regions} Given a teacher feature, $\mathcal{F}^{\mathcal{T}}_{\text{enc}}(x)$, its attention map can be denoted as $\mathcal{A}(\mathcal{F}^{\mathcal{T}}_{\text{enc}})(x)$. Then, we select $K$ regions with the $K$ largest attention values as the crucial regions in this image. Denote the index set of regions as  
  $P_K$, then the feature of the crucial regions can be formulated as $\mathcal{G}(x)=\text{stack}(\{\mathcal{F}_{\text{enc}}[:,i]\}) \in \mathbb{R}^{c\times K}$ where i $\in$ $P_k$. Denote its index set as $S'=\{1,2,...,K\}$, then our region-aware knowledge distillation can be formulated as 
 
  \begin{equation}
  \label{equ:region_aware}
  \begin{aligned}
      &\mathcal{L}_{\text{\texttt{ReKo}}}(\mathcal{X}, \mathcal{G}^\mathcal{S}, \mathcal{G}^\mathcal{T}) \\ &= \mathbb{E}_{x \sim \mathcal{X}}\mathop{\sum}^{K}_{i=1} \mathcal{L}_{\text{InfoNCE}}(\underbrace{\mathcal{G}^{\mathcal{S}}_{}(x)[:,i]}_{\text{Query}},\underbrace{\mathcal{G}^{\mathcal{T}}_{}(x)[:,i]}_{\text{Positive Key}},\underbrace{
      \{ \mathcal{G}^\mathcal{T}[:,j]~|~ j\in S' , j\neq i \})}_{\text{Negative Keys}},
      \end{aligned}
  \end{equation}It is observed that the main difference between $\mathcal{L}_{\text{RegionDis}}$ and $\mathcal{L}_{\text{\texttt{ReKo}}}$ is that $\mathcal{L}_{\text{\texttt{ReKo}}}$ applies knowledge distillation only to the $K$ crucial regions found by $\mathcal{A}$ instead of all the regions. Based on the above formulation, we can introduce the overall training loss of students as 
  \vspace{-0.05cm}
  \begin{equation}
      \mathcal{L}_{\text{Student}} = 
  \alpha \cdot\mathcal{L}_{\text{ReKo}} + 
      \mathcal{L}_{\text{Origin}},
  \end{equation}
  where $\mathcal{L}_{\text{Origin}}$ is the original training loss of image-to-image translation models. For instance, in Pix2Pix, $L_{\text{Origin}}$ indicates the adversarial learning loss and the mean square loss between the ground truth and the generated images. We do not introduce it here in detail since it has no direct influence with our method. 
   $\alpha$ is a hyper-parameter to balance the two loss functions.
   Sensitively study in Section~\ref{fig:sensitivity} shows that our method is not sensitive to its value.
  
\vspace{-0.2cm}
  
  \subsubsection{Trick - Freezing the Projection Head}
  As pointed out by previous works in contrastive learning, the architecture and training methods of the projection heads have a significant influence on the performance of contrastive learning\cite{simclr,chen2020exploring}. Recently, MoCov3 shows that the training stability of contrastive learning on vision transformers can be improved by using a fixed random projection head\cite{chen2021empirical}. 
  Similarly, in \texttt{ReKo}, we propose to initialize the projection head randomly and freeze its weights during the whole training period. Experimental results show that this trick can stabilize student training and lead to better performance.

  \section{Experiments}

  \subsection{Experiment Setting}

  \subsubsection{Models and Datasets} 
  We evaluate \texttt{ReKo} on three image-to-image translation models including CycleGAN~\cite{cyclegan} for unpaired image-to-image translation,  Pix2Pix~\cite{pix2pix} and Pix2PixHD~\cite{pix2pixHD} for paired image-to-image translation. 
  Three datasets including Horse$\leftrightarrow$Zebra, Edge$\rightarrow$Shoe and Cityscapes are utilized for quantitative evaluation. Horse$\leftrightarrow$Zebra
  is an unpaired image-to-image translation dataset which translates images of horses to zebras and vice versa. Edge$\rightarrow$Shoe is a paired image-to-image dataset which maps the edges of shoes to their natural images.
  Cityscapes is a dataset which translates the semantic segmentation result to its origin image~\cite{Cordts2016Cityscapes}. Besides, we also conduct qualitative experiments on Facades, Maps, Summer$\leftrightarrow$Winter, Apple$\leftrightarrow$Orange, Photo$\leftrightarrow$Monet and Photo$\leftrightarrow$Vangogh.
  In all the experiments, students have the same architecture and depth with their teachers except for fewer channels. 
  
  \begin{figure*}[h!]
      \centering
      \includegraphics[width=12cm]{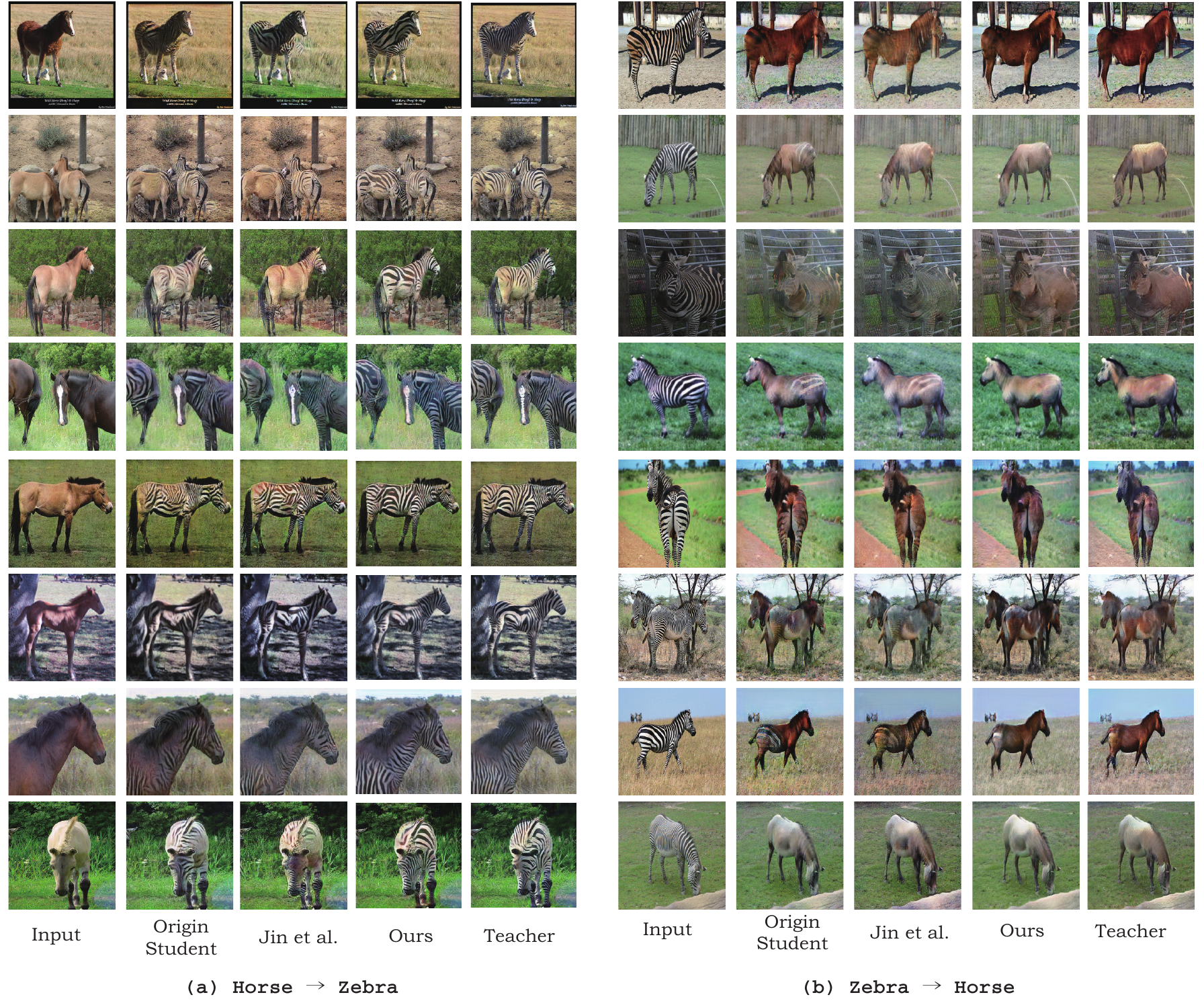}
      \caption{Qualitative results on Horse$\rightarrow$Zebra and Zebra$\rightarrow$Horse. The students are 15.81$\times$ compressed CycleGAN.
      }
      \vspace{-0.4cm}
      \label{fig:qualitative-eps-converted-to.pdf}
  \end{figure*}
 
\begin{table*}[h!]  \caption{\label{tab:quan2} Experimental results on unpaired image-to-image translation on Horse$\rightarrow$Zebra and Zebra$\rightarrow$Horse with CycleGAN.  $\Delta$ indicates the performance improvements (FID decrements) compared with the origin student. 
  Each result is averaged from 8 trials. \textbf{A lower FID is better}.
  }
    \vspace{-0.1cm}
  \begin{center}
  \scriptsize
          \setlength{\tabcolsep}{1.2mm}{
     \begin{tabular}{l l l l l c l }

     \toprule 
     \multirow{2}{*}{Dataset} &\multirow{2}{*}{\#Params~(M)}& \multirow{2}{*}{FLOPs~(G)}&\multirow{2}{*}{Method} & \multicolumn{2}{c}{Metric}\\
     \cmidrule{5-6} 
     &&&&FID$\downarrow$&$\Delta \uparrow$\\
     
     \midrule\multirow{23}{*}{Horse$\rightarrow$Zebrea}&11.38&49.64&Teacher&61.34$\pm$4.35&--\\ \cmidrule{2-6}
     &\multirow{10}{*}{0.72 (15.81$\times$)}&\multirow{10}{*}{3.35 (14.82$\times$)}&Origin Student&85.04$\pm$6.88&--\\
      &&&Hinton~\emph{et al.}~\cite{distill_hinton}&84.08$\pm$3.78&0.96\\
      &&&Zagoruyko~\emph{et al.}~\cite{attentiondistillation}&81.24$\pm$2.01&3.80\\
 
      &&&Li and Lin~\emph{et al.}~\cite{gan_compress}&83.97$\pm$5.01&1.07\\
      &&&Li and Jiang~\emph{et al.}~\cite{spkd_gan}&81.74$\pm$4.65&3.30\\
      &&&Jin~\emph{et al.}~\cite{teacher_do_more_gankd}&82.37$\pm$8.56&2.67\\
      &&&Ahn~\emph{et al.}~\cite{kd_variational}&82.91$\pm$2.41&2.13\\
      &&&Ren~\emph{et al.}~\cite{omgd}&77.31$\pm$6.41&7.73\\
      &&&Li~\emph{et al.}~\cite{revisit_discriminator}&79.29$\pm$7.31&5.75\\
 
      &&&\textbf{Ours}\cellcolor{mygray}&\cellcolor{mygray}\textbf{71.21$\pm$61.7}&\cellcolor{mygray}\textbf{13.83}\\

      \cmidrule{2-6}
      &\multirow{10}{*}{ 1.61~(7.08$\times$)}&\multirow{10}{*}{ 7.29~(6.80$\times$)}&Origin Student&70.54$\pm$9.63&--\\
      &&&Hinton~\emph{et al.}~\cite{distill_hinton}&70.35$\pm$3.27&0.18\\
      &&&Zagoruyko~\emph{et al.}~\cite{attentiondistillation}&67.51$\pm$4.57&3.03\\
 
      &&&Li and Lin~\emph{et al.}~\cite{gan_compress}&68.58$\pm$4.31&1.96\\
      &&&Li and Jiang~\emph{et al.}~\cite{spkd_gan}&68.94$\pm$2.98&1.60\\
      &&&Jin~\emph{et al.}~\cite{teacher_do_more_gankd}&67.31$\pm$3.01&3.23\\
      &&&Ahn~\emph{et al.}~\cite{kd_variational}&69.32$\pm$5.89&1.22\\
    &&&Ren~\emph{et al.}~\cite{omgd}&64.78$\pm$5.21&5.76\\
      &&&Li~\emph{et al.}~\cite{revisit_discriminator}&66.85$\pm$6.17&3.69\\
      &&&\cellcolor{mygray}\textbf{Ours}&\cellcolor{mygray}\textbf{60.01$\pm$5.22}&\cellcolor{mygray}\textbf{10.53}\\
      \midrule
      
\multirow{23}{*}{Zebra$\rightarrow$Horse}&11.38&49.64&Teacher&138.07$\pm$4.01&--\\  \cmidrule{2-6}
&\multirow{10}{*}{0.72 (15.81$\times$)}&\multirow{10}{*}{3.35 (14.82$\times$)}&Origin Student&152.67$\pm$5.07&--\\
&&&Hinton~\emph{et al.}~\cite{distill_hinton}&148.64$\pm$1.62&4.03\\
&&&Zagoruyko~\emph{et al.}~\cite{attentiondistillation}&148.92$\pm$1.20&3.75\\
&&&Li and Lin~\emph{et al.}~\cite{gan_compress}&151.32$\pm$2.31&1.35\\
&&&Li and Jiang~\emph{et al.}~\cite{spkd_gan}&151.09$\pm$3.67&1.58\\
&&&Jin~\emph{et al.}~\cite{teacher_do_more_gankd}&149.73$\pm$3.94&2.94\\
&&&Ahn~\emph{et al.}~\cite{kd_variational}&150.31$\pm$3.55&2.36\\
&&&Ren~\emph{et al.}~\cite{omgd}&146.52$\pm$3.17&6.15\\
      &&&Li~\emph{et al.}~\cite{revisit_discriminator}&147.27$\pm$4.13&5.40\\
&&&\cellcolor{mygray}\textbf{Ours}&\cellcolor{mygray}\textbf{142.58$\pm$4.27}&\cellcolor{mygray}\textbf{10.09}\\
\cmidrule{2-6}
&\multirow{10}{*}{ 1.61~(7.08$\times$)}&\multirow{10}{*}{ 7.29~(6.80$\times$)}&Origin Student&141.86$\pm$1.57&--\\
&&&Hinton~\emph{et al.}~\cite{distill_hinton}&142.03$\pm$1.61&-0.17\\
&&&Zagoruyko~\emph{et al.}~\cite{attentiondistillation}&141.23$\pm$1.27&0.63\\
&&&Li and Lin~\emph{et al.}~\cite{gan_compress}&141.32$\pm$1.27&0.54\\
&&&Li and Jiang~\emph{et al.}~\cite{spkd_gan}&141.16$\pm$1.31&0.70\\
&&&Jin~\emph{et al.}~\cite{teacher_do_more_gankd}&140.98$\pm$1.41&0.88\\
&&&Ahn~\emph{et al.}~\cite{kd_variational}&141.50$\pm$2.51&0.36\\
 &&&Ren~\emph{et al.}~\cite{omgd}&140.58$\pm$1.73&1.28\\
      &&&Li~\emph{et al.}~\cite{revisit_discriminator}&141.15$\pm$1.57&0.71\\
&&&\textbf{Ours}\cellcolor{mygray}&\cellcolor{mygray}\textbf{137.03$\pm$3.03}&\cellcolor{mygray}\textbf{4.83}\\ 

  \bottomrule
  \end{tabular}}

    \vspace{-0.8cm}
  \end{center}
  \end{table*}

\begin{table*}[h!]

\caption{\label{tab:quan1} Experimental results on paired image-to-image translation on Edge$\rightarrow$Shoe with Pix2Pix and Pix2PixHD. $\Delta$ indicates the performance improvements compared with the origin student.
  Each result is averaged from 8 trials. \textbf{A lower FID is better}.
  }
\vspace{-0.15cm}
\begin{center}
\scriptsize
        \setlength{\tabcolsep}{1.2mm}{
   \begin{tabular}{l c c l l c c }
   
   \toprule
    \multirow{2}{*}{Models}&\multirow{2}{*}{\#Params~(M)}& \multirow{2}{*}{FLOPs~(G)}&\multirow{2}{*}{Method} & \multicolumn{2}{c}{Metric}\\
   \cmidrule{5-6}
   &&&&FID$\downarrow$&$\Delta \uparrow$\\
   
\midrule
\multirow{13}{*}{Pix2Pix}&54.41&6.06&Teacher&59.70$\pm$0.91&--\\
 
 \cmidrule{2-6}
 &\multirow{12}{*}{13.61 (4.00$\times$)}&\multirow{12}{*}{1.56 (3.88$\times$)}&Origin Student&85.06$\pm$0.98&--\\
 &&&Hinton~\emph{et al.}~\cite{distill_hinton}&86.97$\pm$3.49&-1.91\\
 &&&Zagoruyko~\emph{et al.}~\cite{attentiondistillation}&84.25$\pm$2.08&0.81\\
 &&&Li and Lin~\emph{et al.}~\cite{gan_compress}&83.63$\pm$3.12&1.43\\
 &&&Li and Jiang~\emph{et al.}~\cite{spkd_gan}&84.01$\pm$2.31&1.05\\
 &&&Jin~\emph{et al.}~\cite{teacher_do_more_gankd}&84.39$\pm$3.62&0.67\\
 &&&Ahn~\emph{et al.}~\cite{kd_variational}&84.92$\pm$0.78&0.14\\
 &&&Ren~\emph{et al.}~\cite{omgd}&80.31$\pm$2.59&4.75\\
      &&&Li~\emph{et al.}~\cite{revisit_discriminator}&81.24$\pm$3.74&3.82\\
 &&&\textbf{Ours}\cellcolor{mygray}&\cellcolor{mygray}\textbf{77.69$\pm$3.14}&\cellcolor{mygray}\textbf{7.37}\\
 \cmidrule{4-6}
 &&&Our + Ren~\emph{et al.}~\cite{omgd}&74.24$\pm$2.48&10.85\\
      &&&Our + Li~\emph{et al.}~\cite{revisit_discriminator}&75.21$\pm$3.15&9.85\\
 
 \midrule
     
 \multirow{13}{*}{Pix2PixHD}&45.59&48.36&Teacher&41.59$\pm$0.42&--\\\cmidrule{2-6}
 &\multirow{12}{*}{1.61 (28.23$\times$)}&\multirow{12}{*}{ 1.89~(25.59$\times$)}&Origin Student&44.64$\pm$0.54&--\\
 &&&Hinton~\emph{et al.}~\cite{distill_hinton}&45.31$\pm$0.63&-0.67\\
 &&&Zagoruyko~\emph{et al.}~\cite{attentiondistillation}&44.21$\pm$0.72&0.43\\
 &&&Li and Lin~\emph{et al.}~\cite{gan_compress}&44.03$\pm$0.41&0.61\\
 &&&Li and Jiang~\emph{et al.}~\cite{spkd_gan}&43.90$\pm$0.36&1.28\\
 &&&Jin~\emph{et al.}~\cite{teacher_do_more_gankd}&43.97$\pm$0.17&1.21\\
 &&&Ahn~\emph{et al.}~\cite{kd_variational}&44.53$\pm$0.48&0.11\\
 &&&Ren~\emph{et al.}~\cite{omgd}&42.98$\pm$0.34&1.66\\
      &&&Li~\emph{et al.}~\cite{revisit_discriminator}&43.21$\pm$0.35&1.43\\
 &&&\textbf{Ours}\cellcolor{mygray}&\cellcolor{mygray}\textbf{42.31$\pm$0.17}&\cellcolor{mygray}\textbf{2.33}\\ 
 \cmidrule{4-6}
  &&&Our + Ren~\emph{et al.}~\cite{omgd}&41.25$\pm$0.54&3.39\\
      &&&Our + Li~\emph{et al.}~\cite{revisit_discriminator}&41.88$\pm$0.53&2.76\\
\bottomrule
\end{tabular}}
\vspace{-0.5cm}
\end{center}
\end{table*}

\begin{table}[h!]
  \caption{\label{tab:cityscapes} Experimental results on Cityscapes with Pix2Pix. $\Delta$ indicates the performance improvements compared with the origin student.
  Each experiment is averaged from 8 trials. \textbf{A higher mIoU is better}.}
\vspace{-0.15cm}
  \begin{center}
  \scriptsize
          \setlength{\tabcolsep}{4.0mm}{
     \begin{tabular}{ cc l c c c c }
     
     \toprule \multirow{2}{*}{\#Params~(M)}& \multirow{2}{*}{FLOPs~(G)}&\multirow{2}{*}{Method} & \multicolumn{2}{c}{Metric}\\
      \cmidrule{4-5}
     &&&mIoU$\uparrow$&$\Delta \uparrow$\\
     \midrule
     54.41&96.97&Teacher&46.51$_{\pm0.32}$&--\\
      
      \midrule
      \multirow{8}{*}{13.61 \tabincell{c}{$_{4.00\times}$}}&\multirow{8}{*}{\tabincell{c}{24.90 $_{3.88\times}$}}&Origin Student&41.35$_{\pm0.22}$&--\\
      &&Hinton~\emph{et al.}~\cite{distill_hinton}&40.49$_{\pm0.41}$&-0.86\\
      &&Zagoruyko~\emph{et al.}~\cite{attentiondistillation}&40.17$_{\pm0.36}$&-1.18\\
      &&Li and Lin~\emph{et al.}~\cite{gan_compress}&41.52$_{\pm0.34}$&0.17\\
      &&Li and Jiang~\emph{et al.}~\cite{spkd_gan}&41.77$_{\pm0.30}$&0.42\\
      &&Jin~\emph{et al.}~\cite{teacher_do_more_gankd}&41.29$_{\pm0.51}$&-0.06\\
       &&Ahn~\emph{et al.}~\cite{kd_variational}&41.88$_{\pm0.45}$&0.53\\
       &&Ren~\emph{et al.}~\cite{omgd}&42.31$\pm$0.31&0.96\\
      &&Li~\emph{et al.}~\cite{revisit_discriminator}&41.75$\pm$0.42&0.40\\
      && Ours\cellcolor{mygray}&\cellcolor{mygray}\textbf{42.81$_{\mathbf{\pm0.25}}$}&\cellcolor{mygray}\textbf{1.46}\\
     \bottomrule
  \end{tabular}}

\vspace{-0.5cm}
  \end{center}
  \end{table}

  \subsubsection{Comparison Methods}
  The students in our experiments have the same neural network depth but fewer channels compared with their teachers. Eight knowledge distillation methods are adopted for comparison. For Hinton  knowledge distillation, we adopt it for image-to-image translation by replacing the KL divergence loss with $\mathcal{L}_1$-norm loss between the generated images of students and teachers.
  Note that some of comparison methods (Li~\emph{et al.}~\cite{gan_compress}, Jin~\emph{et al.} ~\cite{teacher_do_more_gankd} and Li~\emph{et al.}~\cite{revisit_discriminator}) includes both neural network pruning and knowledge distillation.  For a fair comparison, we only compare \texttt{ReKo} with their knowledge distillation parts. 
  
  \vspace{-0.1cm}

  \subsubsection{Evaluation Settings}
  On Cityscapes, we follow the previous works to evaluate the quality of generated images with the mIoU score of a pre-trained segmentation model. 
  On the other datasets, 
  \emph{Fr\'echet Inception Distance (FID)}, which measures the distance between the distribution of features extracted from the real and the synthetic images, is utilized as the metric for all the experiments. A lower FID and a higher mIoU indicate that the synthetic images have better quality. 
  Moreover, to obtain more reliable results, we run 8 trials for each experiment and report their average and standard deviation with the following form: {FID\footnotesize}$_{\pm\text{std. dev.}}$ / {mIoU\footnotesize}$_{\pm\text{std. dev.}}$ instead of reporting the lowest FID. 
  

  \begin{figure*}[t!]
    \centering
    \includegraphics[width=12cm]{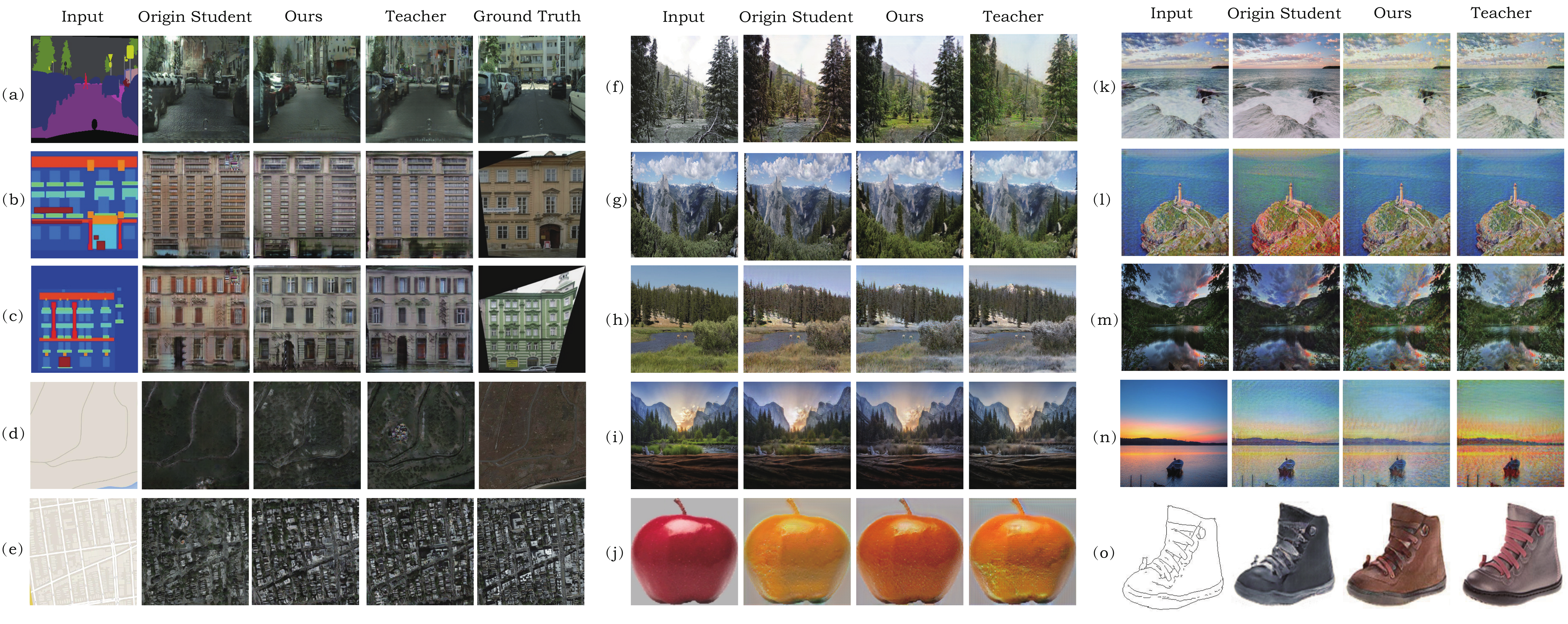}

    \caption{Qualitative experiments on the other datasets: Cityscapes (a), Facades (b-c),  Maps (d-e), Edge$\rightarrow$Shoe (o) with Pix2Pix for paired image-to-image translation and Winter $\rightarrow$Summer (f-g),  Summer$\rightarrow$Winter (h-i),  Apple$\rightarrow$Orange (j), Photo$\rightarrow$Monet (k-l), Photo$\rightarrow$Vangogh (m-n) for unpaired image-to-image translation. Pix2Pix students on Cityscapes and the other datasets are 4.00$\times$ and 28.32$\times$ times compressed, respectively. CycleGAN students are 15.81$\times$ compressed.\label{fig:more} }
\end{figure*}

\subsection{Experimental Result}
  \subsubsection{Quantitative Result} Quantitative results of \texttt{ReKo} compared with eight knowledge distillation methods have been shown in Table~\ref{tab:quan2}, Table~\ref{tab:quan1} and Table~\ref{tab:cityscapes}.
  It is observed that:
  \textbf{(a)} \texttt{ReKo} leads to consistent and significant performance improvements (FID reduction) on various datasets and models. On average, it leads to 9.2 and 4.85 FID reduction on unpaired and paired image-to-image translation tasks, respectively.
  \textbf{(b)} \texttt{ReKo} outperforms the other eight kinds of image-to-image translation knowledge distillation methods by a large margin. On average, it outperforms the second-best method by 3.61 FID.
  \textbf{(c)} Not all the knowledge distillation methods work well on GAN for image-to-image translation. Directly applying the na\"ive Hinton knowledge distillation~\cite{distill_hinton} leads to limited and sometimes even  negative effects. For example, it leads to 1.91 FID increment on the Pix2Pix student for Edge$\rightarrow$Shoe.
  \textbf{(d)} 
  Compared with paired image-to-image translation, there are more performance improvements on unpaired image-to-image translation with \texttt{ReKo}. This is probably caused by the fact that there is less labeled supervision in unpaired image-to-image translation. Thus the knowledge from teachers is more helpful. 
  \textbf{(e)} A high ratio of acceleration and compression can be achieved by replacing the teacher model with the distilled student model. For example, 
  \texttt{ReKo} leads to 7.08$\times$ compression and 6.80$\times$ acceleration on CycleGAN students in terms of parameters and FLOPs.  The compressed students outperform their teachers by 1.33 and 1.04 FID on the tasks of Horse$\rightarrow$Zebra and Zebra$\rightarrow$Horse, respectively. \textbf{(f)} As shown in Table~\ref{tab:quan1}, our method can be utilized with previous methods together to achieve better performance.

  \subsubsection{Qualitative Result}
  Qualitative results on Horse$\rightarrow$Zebra and Zebra$\rightarrow$Horse have been shown in  Figure~\ref{fig:qualitative-eps-converted-to.pdf}. It is observed that the student model trained without knowledge distillation can not always translate the whole body of horses and zebras. In contrast, the student model trained with \texttt{ReKo} does not have this issue. Moreover, on Horse$\rightarrow$Zebra, the student model trained by \texttt{ReKo} sometimes outperforms its teacher on the effect of removing the stripes in zebras.  Figure~\ref{fig:more} shows more qualitative results on the other several tasks. It is observed that \texttt{ReKo} also leads to consistent and  significant image quality improvements on all of them. Specifically, on the tasks which all the image should be translated such as Summer$\rightarrow$Winter, \texttt{ReKo} still leads to significant improvements on the generated images, indicating that \texttt{ReKo} is not limited in only the tasks where the to-be-transformed object is clearly defined.

  \section{Discussion}

  \begin{figure*}[t!]
    \centering
    \includegraphics[width=12cm]{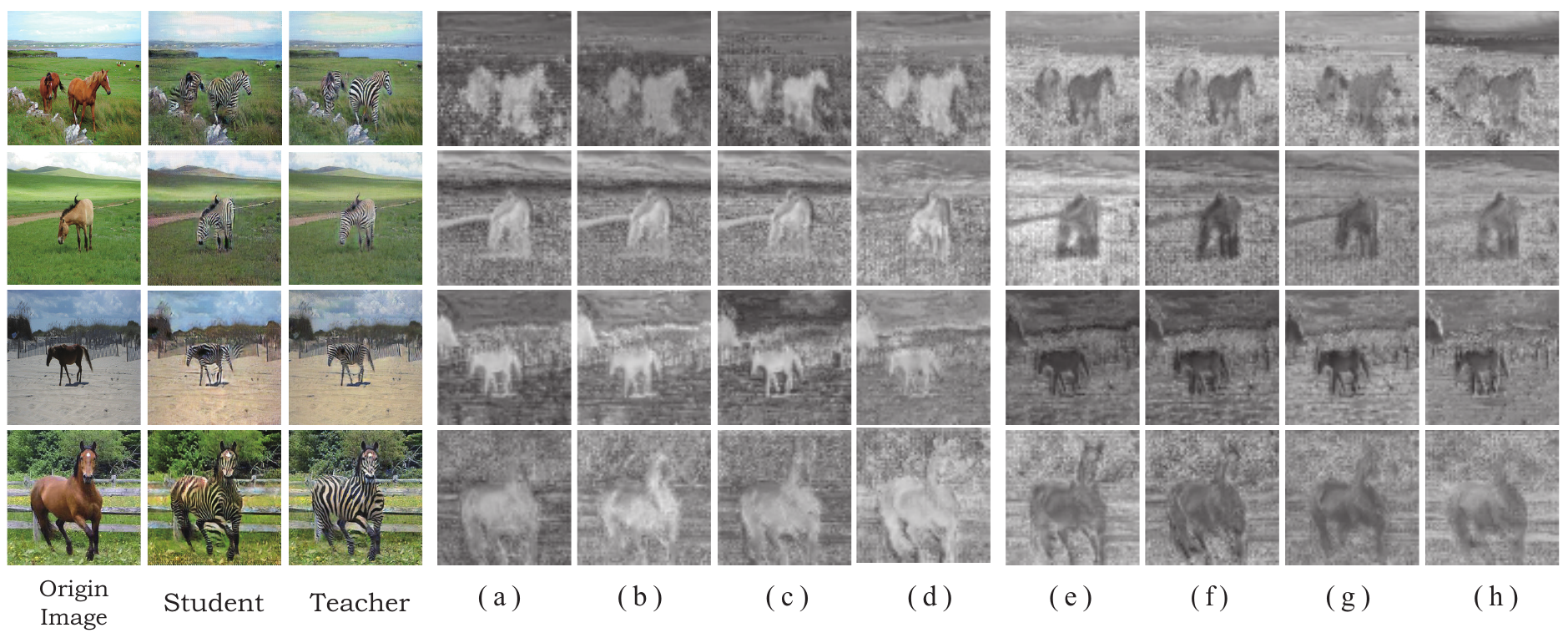}
    \caption{The visualization on the similarity between student features (query) and teachers  features (keys). (a-d) and (e-h) show the similarity between all teacher features and four student features in regions of the horse body and the background, respectively. 
    \label{fig:similarity}}
  \end{figure*}

  \subsection{Visualization on Similarity}
  As shown in Figure~\ref{fig:region_aware} (c), \texttt{ReKo} distills teacher knowledge to the student by improving their feature similarity on the same region. In this subsection, these similarity learned by the student is visualized in Figure~\ref{fig:similarity}. Each gray-scale figure shows the similarity between teacher features in all the regions and student features in one region. 
  In Figure~\ref{fig:similarity} (a-d) and (e-h), the student features are chosen from regions in the horse body and the background, respectively. 
  It is observed that student features in the horse body tend to have higher similarity with teacher features in the hose body and vice versa. These observations demonstrate that student and teacher feature similarity in the same region is successfully increased by optimizing the knowledge distillation loss ($\mathcal{L}_{\text{ReKo}}$), and thus teacher knowledge has been effectively distilled to the students.
  
  \subsection{Ablation Study}

\begin{figure*}[t]
\vspace{-0.3cm}
    \makeatletter\def\@captype{figure}\makeatother
    \begin{minipage}{.47\textwidth}
      \includegraphics[width=6cm]{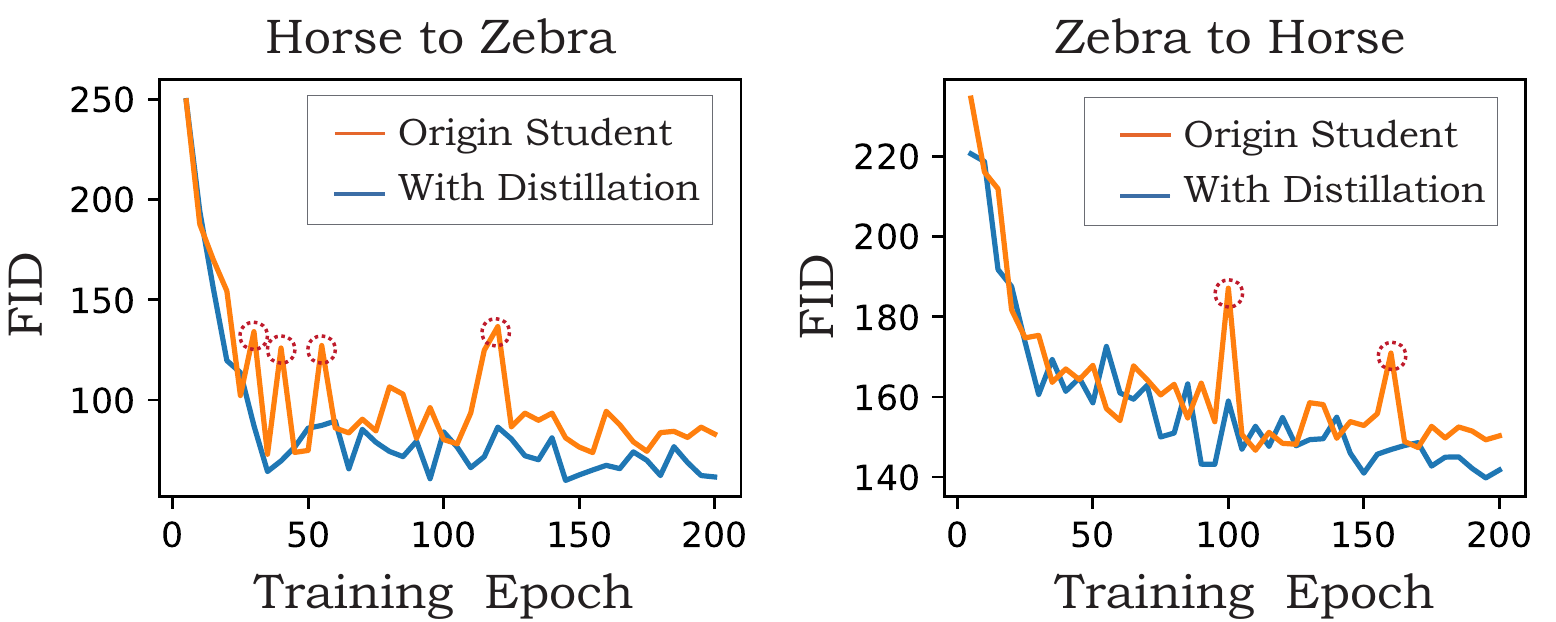}
      \vspace{-0.6cm}
    \caption{\label{fig:stabilize}
    FID curves of two CycleGAN student trained with and without knowledge distillation during the training period.}
  
    \end{minipage}
    \begin{minipage}{.02\textwidth}
        ~~
    \end{minipage}
    \begin{minipage}{.02\textwidth}
      ~~
  \end{minipage}
    \makeatletter\def\@captype{table}\makeatother
    \begin{minipage}{.43\textwidth}

      \footnotesize\caption{Ablation studies on CycleGAN for Horse$\rightarrow$Zebra. \texttt{\textbf{CR}}:~\textbf{C}rucial \textbf{R}egions. \texttt{\textbf{CD}}:~Patch-wise \textbf{C}ontrastive learning for knowledge \textbf{D}istillation. \textbf{Lower FID is better}.  \label{tab:ablation}}
      \vspace{0.2cm}
      \begin{center}
        \setlength{\tabcolsep}{1.0mm}{\begin{tabular}{c|ccccccc}
    
          \toprule
      
          {\texttt{\textbf{CR}}}            &$\times$   &$\times$ &$\checkmark$      &$\checkmark$        \\
              {\texttt{\textbf{CD}}}   &$\times$   &$\checkmark$ &$\times$   &$\checkmark$                                                           \\
          \midrule
          FID ($\downarrow$)&70.54&65.53&67.31&60.01\\
          \bottomrule
          \end{tabular}}%
          
          \end{center}
    \end{minipage}
    \end{figure*}

  Ablation studies on the individual effectiveness of \emph{distilling only the \textbf{C}rucial \textbf{R}egions }(\texttt{CR}) and \emph{patch-wise \textbf{C}ontrastive learning for knowledge \textbf{D}istillation} (\texttt{CD}) have been shown in Table~\ref{tab:ablation}.  Note that when \texttt{CD} is disabled but \texttt{CR} is used, we directly minimize the $L_2$-norm distance between teacher and student features in crucla regions for optimization.
  It is observed that 5.01 and 3.23 FID reduction can be obtained by using only \texttt{CD} and \texttt{CR}, respectively. Besides, combining the two methods further leads to a 5.52 FID reduction. These observations indicate that both of the two modules are effective, and their merits are orthogonal.

  \subsection{KD Stabilizes GAN Training}
  The training of GAN is usually not stable due to their complex network architectures and loss functions. In this paper, we find that the proposed knowledge distillation can alleviate this problem.
  Figure~\ref{fig:stabilize} shows the FID curves of CycleGAN students in different training epochs on Horse$\rightarrow$Zebra and Zebra$\rightarrow$Horse. It is observed that: \textbf{(a)} Both the training of students with and without knowledge distillation are stable in the first several epochs. \textbf{(b)} After the early epochs,
  the training of the student without knowledge distillation becomes unstable and sometimes collapses (marked with circles). In contrast, the distilled student is consistently stable during the whole training period. Its FID undulations are much smaller than the student trained without knowledge distillation.
  
  \subsection{Find Crucial Regions with other Methods\label{sec:other}}

  In \texttt{ReKo}, the attention of the teacher network is utilized to find the crucial regions in an image. In this section, we further evaluate the performance of \texttt{ReKo} with the following four schemes: (1)\emph{student attention} - localizing crucial regions with attention of the student; (2)\emph{VGG attention} - localizing crucial regions with the attention from a ImageNet pre-trained VGG model; (3)\emph{VGG Grad-CAM} localizing crucial regions with the Grad-CAM result from a ImageNet pretrained VGG respect to the to-be-transformed object (\emph{e.g.} horses and zebras); (4)\emph{salient detection} - localizing crucial regions with unsupervised salient detection. Experiments with 7.08$\times$ compressed CyCleGAN on Horse$\rightarrow$Zebra show that our scheme (with teacher attention) and the above four schemes achieve 60.01, 61.46, 62.91, 63.46 and 65.09 FID, respectively, indicating that teacher attention is the most effective metric in \texttt{ReKo} to localize the crucial regions for knowledge distillation and student attention and Grad-CAM are also two effective solutions.

  \subsection{Sensitivity Study to Hyper-parameters\label{fig:sensitivity}}
\begin{figure}[t!]
    \begin{center}
        \includegraphics[width=12cm]{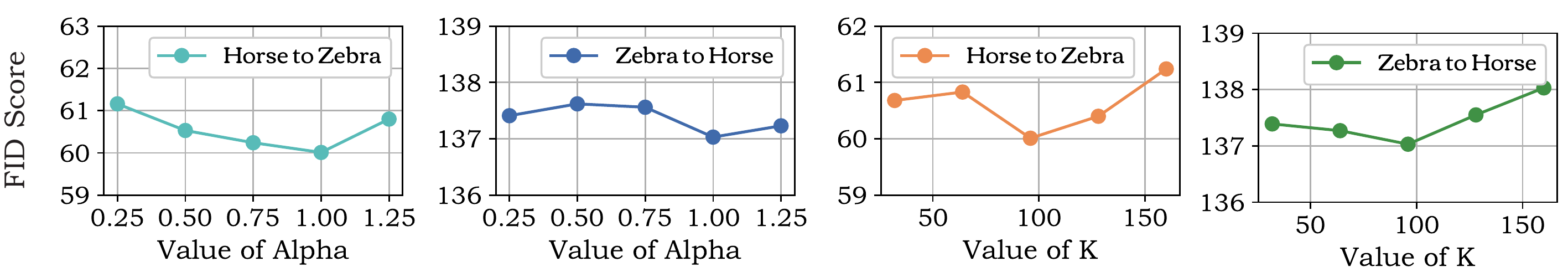}
        \vspace{-0.3cm}
    \end{center}
    \caption{\label{fig:hyper} Sensitivity study to hyper-parameters $\alpha$ and $K$ on Horse$\rightarrow$Zebra and Zebr$\rightarrow$Horse with 14.82$\times$ compressed CycleGAN students. }
    \vspace{-0.2cm}
\end{figure}

  There are mainly two hyper-parameter $\alpha$ and $K$ introduced in our method. $\alpha$ is utilized to balance the magnitude of different loss functions. $K$ is the number of crucial regions selected in an image. 
  Sensitivity studies on Horse$\rightarrow$Zebra and Zebra$\rightarrow$Horse with CycleGAN students have been shown in Figure~\ref{fig:hyper}. It is observed that all our method is not sensitive to the value of hyper-parameters.
  
  \section{Conclusion}

  This paper proposes region-aware knowledge distillation (\texttt{ReKo}) for compressing image-to-image translation models. An attention module is utilized to localize the crucial regions in the to-be-translated images. Then, patch-wise contrastive learning is employed for knowledge distillation, which maximizes the mutual information between the features of students and teachers in the same region. Abundant experiments with eight comparison methods have been conducted to demonstrate the effectiveness of \texttt{ReKo}. On average, there are 9.2 FID and 4.85 FID reduction on unpaired and paired image-to-image translation tasks, respectively. Our 7.08$\times$ compressed and 6.80$\times$ accelerated CycleGAN student outperforms its teacher by 1.33 and 1.04 FID on Horse$\rightarrow$Zebra and Zebra$\rightarrow$Horse respectively. Additionaly, it can be utilized with the other knowledge distillation methods to achieve better performance.
  Moreover, visualization results on the similarity between student and teacher features in different regions show that teacher knowledge has been effectively distilled. Besides, we show that \texttt{ReKo} can stabilize the training of GANs and prevent them from model collapse.

%
%
\bibliographystyle{splncs04}
\bibliography{egbib}
\end{document}